\documentclass{IOS-Book-Article}

\usepackage{mathptmx}
\usepackage{amssymb} % hinzugefügt.
\usepackage{mathtools} % hinzugefügt.
\usepackage{tabularx} %hinzugefügt.
\usepackage{caption} %hinzugefügt.
\usepackage{subfigure} %hinzugefügt.
\usepackage{multicol} %hinzugefügt.

\def\hb{\hbox to 10.7 cm{}}

\begin{document}

\pagestyle{headings}
\def\thepage{}
\newcommand{\DSAA}{DSA$^2$}

\begin{frontmatter}              % The preamble begins here.

%\pretitle{Pretitle}
\title{Simulated Autonomous Driving on Realistic Road Networks\\ using Deep Reinforcement Learning}

\markboth{}{March 2018\hb}
%\subtitle{Subtitle}

\author[A]{Patrick Klose},
\author[B]{Rudolf Mester}
\address[A]{mail@patrickklose.com}
\address[B]{mester@vsi.cs.uni-frankfurt.de}

\begin{abstract}
Using Deep Reinforcement Learning (DRL) can be a promising approach to handle various tasks in the field of (simulated) autonomous driving.
However, recent publications mainly consider learning in unusual driving environments.
This paper presents Driving School for Autonomous Agents (\DSAA), a software for validating DRL algorithms in more usual driving environments based on artificial and realistic road networks.
We also present the results of applying \DSAA\:for handling the task of driving on a straight road while regulating the velocity of one vehicle according to different speed limits.
\end{abstract}

\end{frontmatter}
\markboth{March 2018\hb}{March 2018\hb}

\section{Introduction}
In the field of autonomous driving, a vehicle can be well thought of as an autonomous agent acting in a complex environment. 
Hence, the particular machine learning paradigma called Reinforcement Learning (RL), which can be broadly defined as the \textit{computational approach to learning from interaction with an environment}, could be a promising idea to make progress in this field \cite{reinforce-learn}.
Indeed, some authors recently showed on the basis of simulations that especially the field of Deep Reinforcement Learning (DRL; the combination of Deep Learning and RL) can be a promising approach to create autonomous agents capable of learning to drive in yet simplified environments \cite{drl-framework, drl-lane-keeping, drl-stanford}.\\\\
All the authors of the mentioned studies validated their applied DRL algorithms in a racing car simulator called \textit{The Open Racing Car Simulator (TORCS)} \cite{software:torcs}, which does not provide a \textit{usual} driving environment. 
Since this drawback is precipitated through the software itself, it does not vanish, even if the algorithms are becoming more sophisticated.
To make further progress in the application of RL to autonomous driving, we therefore developed a simulation software called \textit{Driving School for Autonomous Agents (\DSAA)}\footnote{The software can be downloaded at http://www.patrickklose.com.} for validating DRL algorithms in more usual driving environments based on artificial and realistic road networks.

\section{Driving School for Autonomous Agents (\DSAA)}
\subsection{Overview} \label{s:o}
\begin{figure}[ht]
\centering
\frame{
\includegraphics[scale=0.65]{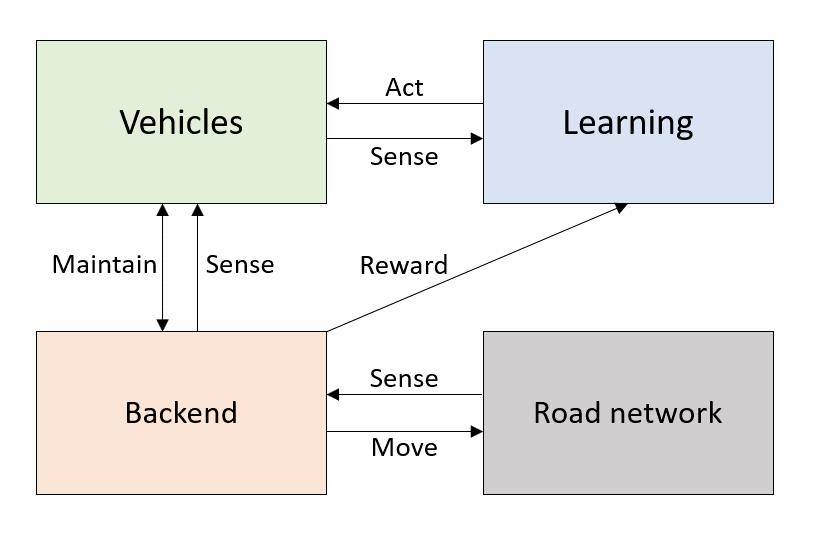}}
	\caption{\DSAA\:modules}
	\label{f:modules}
\end{figure}

\DSAA\:provides a mean to validate DRL algorithms facing various tasks in more usual driving environments based on artificial and realistic road networks.
The software mainly consists of four modules which are depicted in figure \ref{f:modules}.
\\
While the simulation is running, the backend module continuously let the vehicles sense their driving environment as well as maintain and move them (according to their velocity) along their paths.
The current velocity of a vehicle is determined based on a physics model, where the learning module of the software (the DRL algorithm in place) controls throttle and brake based on the representation of the vehicles' current state (further: state).
The learning module in turn is receiving rewards for its decisions from the backend module allowing it to learn to fulfill a previously defined task.\\\\
The physics model of \DSAA\:is focusing on the longitudinal dynamics of the vehicles because this already pose - together with road networks - novel and complex driving tasks for recent DRL algorithms.
Hence, the software is far from providing an absolutely \textit{realistic} driving environment, but instead a potential starting point for changing into more \textit{usual} driving environments.
Finally, worth mentioning is that \DSAA\:is developed to focus on evaluating DRL algorithms of a special kind, i.e., algorithms which are based on the actor-critic architecture.\footnote{Please see \cite{reinforce-learn} for more information about the actor-critic architecture.}\\\\
The following sections provide information about how the software can be used and how the most important parts of \DSAA\:are implemented.
However, please note that the learning module of \DSAA\:is dependent on the particular DRL algorithm in place such that no comprehensive and always valid explanation can be given in this study.
Nevertheless, \DSAA\:provides a framework which eases the implementation of a particular DRL algorithm inside the code.

\subsection{Features}
\subsubsection{Graphical User Interface (GUI)}
\begin{figure}[ht]
\centering
\frame{
  \includegraphics[scale=0.317]{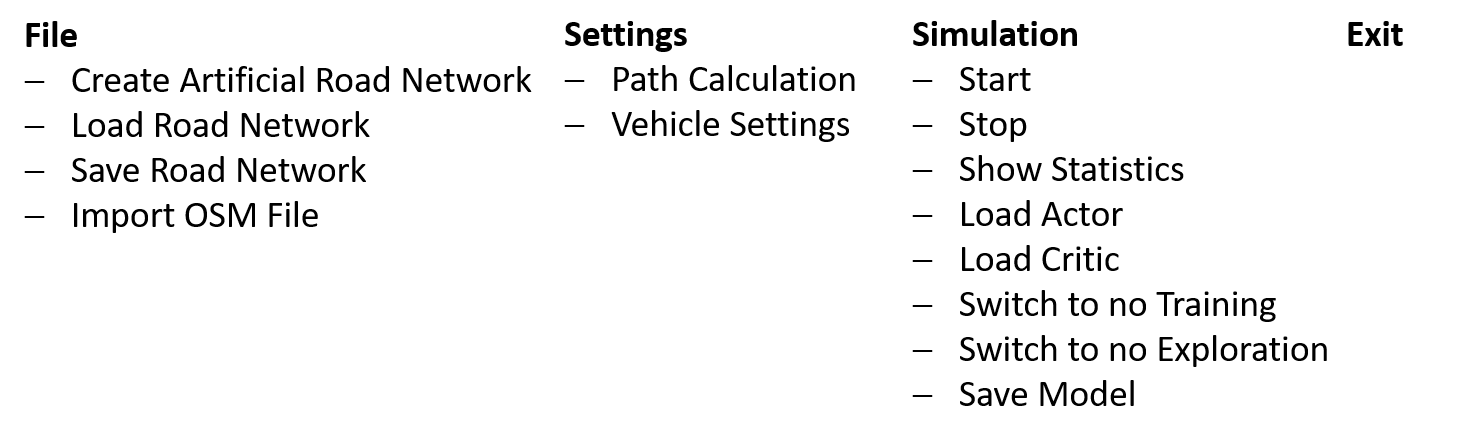}}
	\caption{Menu structure of the GUI}
	\label{f:menu}
\end{figure}

In the field of machine learning, where the development process of an algorithm often involves some form of trial-and-error, a suitable visualization of the algorithms' behavior and of the system as a whole can greatly aid the development process \cite{reinforce-learn, deep-learn, inf-vis}.
Therefore, \DSAA\:provides a Graphical User Interface (GUI) with which the user of the software (further: user) can directly observe the driving and learning of the vehicles as well as can interact with the software in a comfortable manner.\\\\
Starting at the top of the window, the GUI provides a menu (see figure \ref{f:menu} for an overview of its structure) to call the different features of the software.
Below this menu, the GUI shows the loaded road network in a two-dimensional bird's-eye perspective, if a road network is already loaded as well as the placed vehicles, if vehicles are already placed.
Furthermore, if a road network is already loaded, the user can use its mouse wheel to zoom into and out of it.
\\
During the simulation, the positions of the vehicles are updated accordingly and their colors represent their respective average reward over the last 1,000 time steps of the simulation.
The colors range from dark red representing the minimum, to dark green representing the maximum reward ever occured since the start of the simulation.
This directly shows - in an informal manner -  if the DRL algorithm in place is capable of handling the stated task after a certain time period.

\subsubsection{Create Artificial Road Networks}
\begin{figure}[ht]
\centering
  \includegraphics[scale=0.7]{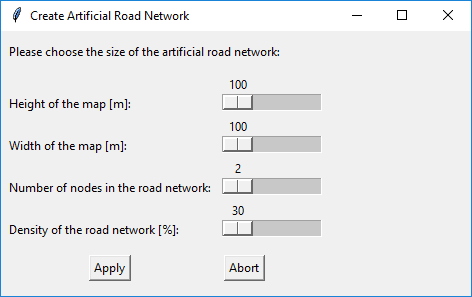}
	\caption{Adjustable properties for artificial road networks}
	\label{f:carn}
\vspace*{-0.02cm}
\end{figure}

This feature provides the possibility to create artificial road networks with the height $[m]$ and width $[m]$ of the corresponding map as well as with the number $n_N \coloneqq |N|$ of nodes in the road network and its density $[\%]$ adjustable by the user (see figure \ref{f:carn}).\\
\\
After the specification of the properties by the user, \DSAA\:firstly uses the desired height and width of the map to create two geographical coordinates $\vec{x}_1$ and $\vec{x}_2$ representing the lower left and upper right point of the map, respectively.
The $n_N$ nodes of the road network are then generated by sampling their geographical coordinates according to $\vec{x}_i \sim \vec{U}(\vec{x}_1, \vec{x}_2)$, where $i \in N$ and $\vec{U}$ representing a two-dimensional uniform distribution over the two-dimensional area spanned by $\vec{x}_1$ and $\vec{x}_2$.
The generation of the edges $E$ (the roads) between the nodes is implemented based on Delaunay triangulation leading to a network formed out of triangles which assures that there are no overlapping roads \cite{algo-geo}.
The density of the road network then represents as a percentage value how many edges $n_E \coloneqq |E|$ there should be compared to the number of edges directly after the application of the Delaunay triangulation (represented by $n_{E'} \coloneqq |E'|$). % vllt tauschen.
To finally achieve the desired density, ($n_{E'} - n_E$) edges are randomly (uniform) removed from the road network.\\\\
After this process, the artificial road network is mapped into a graph data structure and is cleaned as well as enhanced by the routines explained in section \ref{s:dc-de}.

\subsubsection{Load and Save Road Networks}
After the generation of an artificial or after importing a realistic road network (see section \ref{s:osm}), each one has a suitable data structure (graph) for an efficient simulation of the vehicles and also provides additionally (generated) metadata (see sections \ref{s:osm} and \ref{s:dc-de}).
With using the loading and saving feature properly, most of the time-consuming processes which arise during the generation or the import of a road network have to be processed only once and can be bypassed when the same road network is used again.

\subsubsection{Import OSM File} \label{s:osm}
We decided to use data provided by \textit{Open Street Maps (OSM)}\footnote{http://www.openstreetmaps.org} as the basis for the realistic road networks because this data should remain freely available and there are third-party service providers from which individual map-excerpts can be downloaded.
\DSAA\:expects that the OSM data comes in the OSM-XML format which usually can be selected as the desired file format at the chosen provider such that no further preprocessing is necessary. %any
Since OSM is an open source project, the data has to be treated as possibly incomplete and noisy.
Therefore, \DSAA\:tries to extract as much as possible out of the data (see below) and after mapping the realistic road network into a graph data structure, the data is cleaned and enhanced by the routines explained in section \ref{s:dc-de}.\\\\
For parsing an OSM-XML file, \DSAA\:exploits the fact that all the entities in such a file are referenced by an unique id to distinguish them from each other and that each node provides at least its geographical coordinates.
Furthermore, most of the edges provide information about the type of the corresponding road, the speed limit on it and if it is a one-way or a two-way road.
These information are used while parsing the file to decide for each entity if it is eligible to get extracted.
If the entity is a node (denoted here by $i$), then the node as well as its geographical coordinates $\vec{x}_i$ always get extracted and $\vec{x}_i$ is directly saved to the node.
If the entity is an edge (denoted here by $j$), then it is firstly checked, if the type of its corresponding road allows to let vehicles drive on it.
If so, then the edge gets extracted along with further (available) information about it, i.e., the speed limit on that edge (denoted by $v^{max}_j\:[\frac{m}{s}]$) as well as if the edge should be handled as a directed (one-way road) or an undirected one (two-way road).
In the case of a directed edge, the edge gets extracted and $v^{max}_j$ is saved to it (if available).
However, in the case of an undirected edge, two directed edges with opposite directions are generated and $v^{max}_j$ is saved to both (if available).
If there is no information about the possible directions on that road, the corresponding edge is treated as an undirected one and if there is no information about $v^{max}_j$, the corresponding field is left blank.
Finally, if the type of the street does not allow to let vehicles drive on it or is unknown, the corresponding edge not get extracted at all and the parser moves on to the next entity.\\\\
After the process of parsing, the set of nodes $N$ as well as the set of edges $E$ together contain all the relevant entities with all the available and relevant information about them.
This data is then mapped into a graph data structure and is cleaned as well as enhanced by the routines explained in section \ref{s:dc-de}.

\subsubsection{Path Calculation} \label{s:pc}
\begin{figure}[ht]
\centering
  \includegraphics[scale=0.8]{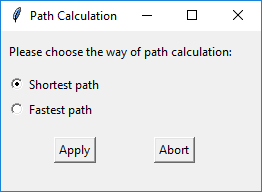}
	\caption{Adjustable options for the path calculation}
	\label{f:pc}
\end{figure}

Each vehicle always has its own path which it follows (from a start node $i\in N$ to a goal node $j\in N$) with $i \neq j$.
Figure \ref{f:pc} shows the two options from which the user can select.
To implement the path calculation for both options, \DSAA\:exploits the fact that a loaded road network is internally represented as a graph such that Dijkstra's algorithm \cite{dijkstra} can be applied.
For Dijkstra's algorithm to work properly, there have to be weights $w_k$ with $w_k>0$ for all the edges $k\in E$ representing a \textit{suitable} form of a distance measure between the two adjacent nodes of an edge \cite{dijkstra}.\\\\
The first option is implemented by using the realistic distance between the two adjacent nodes of an edge as its weight and the second option is implemented by using an adjusted version of these weights to also incorporate the speed limit on a road.
Please see section \ref{s:dc-de} for further details about how these weights for both options (shortest and fastest paths) are calculated.

\subsubsection{Vehicle Settings}
\begin{figure}[ht]
\centering
  \includegraphics[scale=0.7]{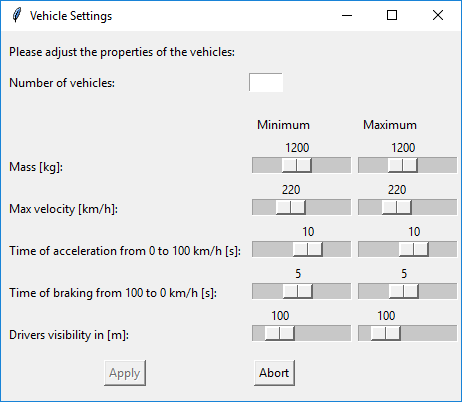}
	\caption{Adjustable properties for the vehicles}
	\label{f:vs}
\end{figure}

In the vehicle settings, the user can specify how many vehicles should be placed onto the road network and which properties these vehicles should have.
As figure \ref{f:vs} shows, the first option (counting from the top) represents the number $n_V \coloneqq |V|$ of vehicles to place, whereby $V$ is the set of all the vehicles after their placement and $1\leq n_V\leq n_N$ must hold.\\
\\
The next five options set the properties of the vehicles and are to be specified as ranges, whereby the particular value for a vehicle is then drawn from a uniform distribution and rounded afterwards.
The specified values act hereby as the lower and upper bound of the corresponding uniform distribution and if for one option the upper and lower bound is equal, the resulting value for this option is deterministic and equal for all the vehicles.\\\\
%Please see section \ref{s:mm}, which explains how the drawn values are internally used to calculate particular parameters for the vehicles, such that they have (at least approximatively) their desired properties.\\\\
After the vehicle settings are specified, $n_V$ vehicles are placed with their particular properties onto the loaded road network.
Furthermore, in the process of placement, each vehicle receives its particular start and goal node, i.e., $i,j\in N$ with $i\neq j$ as well as its path from $i$ to $j$ as explained in section \ref{s:pc}.
Here, $i$ and $j$ are randomly chosen according to $i\sim U(N^-)$ and $j\sim U(N\setminus\{i\})$, where $N^-\subseteq N$ is the set of all the nodes $N$ except for the ones already having a vehicle on it and where $U$ is an one-dimensional uniform distribution over a set expressing that each object in the set get equiprobably selected.
Finally, it is worth mentioning here that the selection of $i$ and $j$ as well as the path calculation are repeated until a path between $i$ and $j$ is found (it is possible that there is no path from $i$ to $j$ due to the existence of one-way roads).

\subsubsection{Show Statistics}
This feature provides a second window on top of the GUI, with which the user can see important statistics about the behavior of a vehicle $i\in V$.
The user can specify the id $i_{id}$ of the vehicle for which statistics should be presented, whereby it holds that $i_{id}\in\{0, 1, ..., n_V-1\}$, i.e., the vehicles are numbered consecutively, starting from zero in an ascending order.\\\\
After specifying $i_{id}$, the user will find the statistics of the vehicle $i$ and they will update accordingly while the simulation is running.
All plots show data for the last 1,000 time steps of the simulation and together comprise the actions taken, the velocities driven, the longitudinal and lateral accelerations experienced and the corresponding rewards earned.

\subsubsection{Load Actor and Critic}
\DSAA\:is focussing on validating DRL algorithms which are based on the actor-critic architecture and this feature provides the possibility to load a previously trained model back into the software.
It can be used to continue the training of a pre-trained model or simply to simulate vehicles based on an already trained model without further training.\\\\
Since the actor-critic architecture is comprised by an actor and a critic, this feature expects two different files to be specified by the user (one for the actor and one for the critic).
Both files have to be in the *.h5 file format and the user has to be sure, which file corresponds to the actor and which one to the critic.
Please see section \ref{s:sm} for further information about the structure of the file names.

\subsubsection{Switch to no Training/no Exploration}
These two features let the user decide if the currently loaded model should be trained further or not and if the software should overlie the calculated actions of the model with some form of exploration which has to be specified upfront.\\\\
In general, deciding when to stop the training of a model has at least two important functions. 
First, early stopping is a well-known generalization technique and the idea is to stop the training at a time step $t$ with $t<\infty$ to avoid the model from overfitting \cite{deep-learn}.
Second, the user may only want to simulate vehicles without training the model and therefore, there has to be anyway a function to turn off the training.\\\\
Exploration is - apart from the previous considerations - usually always a crucial part of the RL framework which can strongly influence the final performance of a trained model \cite{reinforce-learn}.
However, there may be situations where the user wants to turn off the exploration, e.g., in the situation of only simulating vehicles without training the model because in this case, exploration makes no sense.

\subsubsection{Save Model} \label{s:sm}
This feature is implemented to let the user save a model in its current training state such that it can be used or further trained later on.
Again, since \DSAA\:is focussing on validating DRL algorithms which are based on the actor-critic architecture, this feature generates the following four files:
\begin{itemize}
\item the actor as \textless layer-config\textgreater\_\textless date\textgreater\_\textless time\textgreater\_\textless steps\textgreater\_actor.h5
\item the critic as  \textless layer-config\textgreater\_\textless date\textgreater\_\textless time\textgreater\_\textless steps\textgreater\_critic.h5
\item the config file of the actor as \textless layer-config\textgreater\_\textless date\textgreater\_\textless time\textgreater\_\textless steps\textgreater\_actor.txt
\item the config file of the critic as \textless layer-config\textgreater\_\textless date\textgreater\_\textless time\textgreater\_\textless steps\textgreater\_critic.txt
\end{itemize}
Usually, working with the two *.h5 files is sufficient since the model can be completely reconstructed from these.
However, if the user not only needs the model itself but additional information about it, e.g., the detailed architecture of the model in an human-readable form, the two *.txt files can be used.
The file names of the four files also directly provide some information about the models' architecture, the date and time at which the model was saved as well as about the number of time steps that have already passed (the placeholders will be replaced by actual values).

\subsection{Data Cleaning and Enhancement} \label{s:dc-de}
\subsubsection{Data Cleaning}
After generating an artificial or after importing a realistic road network, the resulting graph can have more than one weakly connected component leading to the situation in which a vehicle driving in component one cannot reach component two since they are not connected to each other.\footnote{Please see \cite{networks} for more information about weakly connected components of a graph.}
However, since the idea of simultaneously simulating multiple vehicles is usually to let them interact with each other, we implemented a data cleaning routine which deletes all the weakly connected components of the graph except for the largest one.
Retaining only the largest weakly connected component of the graph then leads to the largest possible road network for simulation and additionally excludes the case of having more than one seperate road network.

\subsubsection{Data Enhancement}
The data enhancement routine augments the graph by additional generated metadata and can be separated into the following four sub-routines.\\\\
\textit{Distance}\\
In both cases, i.e., if an artificial road network is generated or if a realistic road network is imported, there is no distance value between the two adjacent nodes of an edge.
However, these distances are absolutely necessary for a proper simulation of the vehicles as well as for the path calculation as explained in section \ref{s:pc}.
Therefore, \DSAA\:retrospectively calculates all these distances by using the geographical coordinates of the nodes which are in both cases securely available and as the distance measure, the Great Circle Distance (GCD) calculated based on the Haversine formula is used.\footnote{Please see \cite{heavenly} for more information about the calculation of the GCD based on the Haversine formula.}\\\\ %referenz auf section
\textit{Speed Limit}\\
After importing a realistic road network, the value $v^{max}_i$ representing the speed limit on an edge $i\in E$ can be already existent but can also left blank since OSM data does not provide all these values for sure.
Furthermore, after creating an artificial road network, these values are definitely not existent.
Therefore, while the enhancement routine iterates over all the edges in $E$, all the missing values are replaced by a default one.\\\\
\textit{Weights for Fastest Paths}\\
At this point in time, the augmented graph definitely provides for an arbitrarily edge $i\in E$, the speed limit on it $(v^{max}_i)$ as well as the GCD $d^{GCD}_{j, k}\:[m]$ between its two adjacent nodes $j,k\in N$.
This fact is used to calculate additional edge-weights $w_i$ for all the edges with which it is possible to determine a form of fastest paths as already mentioned in section \ref{s:pc}.
For the implementation, we decided to use the following formula which increases with the distance but decreases with the speed limit:
\begin{equation}
w_i \coloneqq \frac {d^{GCD}_{j, k}} {v^{max}_i}
\end{equation}
\textit{Curves}\\
Before the enhancement routine, the graph of an artificial or a realistic road network does not provide information about curves leading from one edge onto another which is a huge simplification.
This leads to the fact, that all the vehicles almost always experience an unusual change in their driving direction when passing a node (with except for their goal nodes) and this in turn can lead to very large and unrealistic lateral accelerations for them.
To solve this problem, \DSAA\:additionally constructs curves onto all relevant pairs of edges which are then used to generate more realistic lateral accelerations.\\
\begin{figure}[ht]
\centering
\frame{
  \includegraphics[scale=0.304]{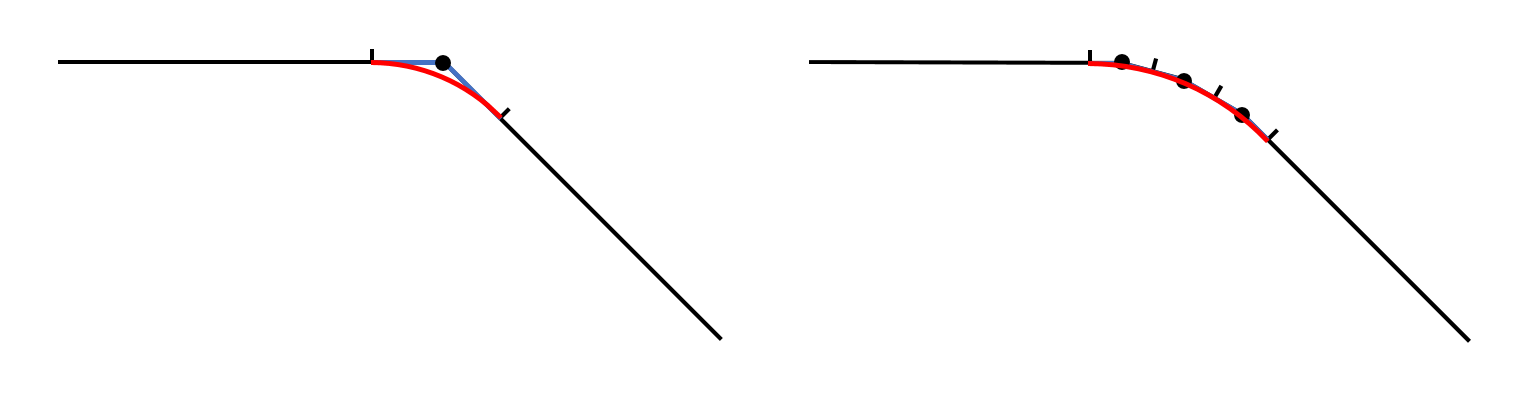}}
	\caption{Two different approximations to a curve}
	\label{f:cca}
\end{figure}\\
To construct all the curves, the sub-routine firstly determines all the relevant pairs by iterating over all the edges in $E$ and using the fact that for an edge $i\in E$ pointing \textit{to} a node $k\in N$, all the edges which point \textit{from} $k$ are the second element of a relevant pair.
With having all the relevant pairs of edges, \DSAA\:then uses basic geometrical principles to construct the curves as segments of circles through three points (see figure \ref{f:cca}; contructed curves are red-colored).
Finally, the sub-routine also accounts for the fact that curves in OSM data are approximated in two different ways, i.e., either by modeling the change in direction based on only one node or by modeling it based on more than one node (see figure \ref{f:cca}).

\subsection{Vehicles}
\subsubsection{The Physics Model} % Respawn has to put anywhere
\DSAA\:restricts the possible actions of a vehicle to (positive) longitudinal acceleration (further: acceleration) and braking, with $A_1=[0,1]$ corresponding to the acceleration actions and $A_2=[-1,0)$ corresponding to the braking actions.
This split between acceleration and braking based on the two disjunct sets $A_1$ and $A_2$ has the advantage that a vehicle cannot accelerate and brake at the same time and that two independent physics models can be used, i.e., the acceleration and the braking model.\\\\
In \DSAA, each model is implemented as an one-dimensional and position-based motion model which takes as the input an action from the learning module and calculates the vehicles' new position based on its current one and the action.
The differences between subsequent positions of a vehicle are then used to correctly move the vehicle along its pre-defined path as well as for generating some of the vehicles' sensors (see section \ref{s:s}).
The acceleration as well as the braking model is implemented as $\vec{z}(t+1) \coloneqq \mathbf{A}_i \cdot \vec{z}(t) + \mathbf{B}_i \cdot a(t+1)$, where $\vec{z}(t)$ describes the position vector at time step $t$, $a(t)\in[-1, 1]$ describes the action at time step $t$ and $i\in\{1, 2\}$ is chosen according to $a(t)$.
The matrices for the acceleration model are defined as
\begin{equation}
\mathbf{A}_1 \coloneqq
\begin{pmatrix}
\frac {\eta \cdot T + 2 \cdot m} {\eta \cdot T + m} & -\frac {m} {\eta \cdot T + m} \\
1 & 0
\end{pmatrix},
\mathbf{B}_1 \coloneqq
\begin{pmatrix}
-\frac {F^{max}_1 \cdot T^3} {-\eta \cdot T^2 - m \cdot T} \\
0
\end{pmatrix},
\end{equation}
where $\eta$ represents the friction coefficient, $T\:[s]$ is the sampling period of the simulation and $m\:[kg]$ as well as $F^{max}_1\:[N]$ are the mass and maximum acceleration force of a particular vehicle, respectively.
The matrices for the braking model are defined as
\begin{equation}
\mathbf{A}_2 \coloneqq
\begin{pmatrix}
\frac {2 \cdot m + \eta \cdot T} {m + \eta \cdot T} & -\frac {m} {m + \eta \cdot T}\\
1 & 0
\end{pmatrix},
\mathbf{B}_2 \coloneqq
\begin{pmatrix}
\frac {T^2 \cdot \hat{g}_0 \cdot \kappa \cdot \tau \cdot m} {m + \eta \cdot T} \\
0
\end{pmatrix},
\end{equation}
where $\hat{g}_0\:[\frac{m}{s^2}]$, $\kappa$ and $\tau$ additionally represent the (approximative) gravitation on earth, the static friction coefficient as well as a correction factor, respectively.
Please note that $\eta$, $m$ , $F^{max}_1$ and $\tau$ are obtained from the vehicles' properties.

\subsubsection{Sensors} \label{s:s}
Sensors are necessary to let the vehicles sense their driving environment and therefore are a part of the driving task.
\DSAA\:mainly focusses on sensors which are usually also used by humans when they drive a vehicle in the real world.
The following list provides all the implemented sensors which \textit{can} be used for defining a vehicles' state.
\begin{multicols}{2}
\begin{itemize}
\item Velocity $[\frac{m}{s}]$
\item Speed limit $[\frac{m}{s}]$
\item Longitudinal acceleration $[\frac{m}{s^2}]$
\item Lateral acceleration $[\frac{m}{s^2}]$
\item Distance to the next curve $[m]$
\item Curve radius at the next curve $[m]$
\item Distance to the next curve at the next intersection $[m]$
\item Distance to the next vehicle on the same path $[m]$
\item Distance to the next vehicle at the next intersection $[m]$
\end{itemize}
\end{multicols}

\section{Application}
In this section, we would like to present the application of \DSAA\:for handling the task of driving on a straight road while regulating the velocity of one vehicle according to different speed limits.
To represent the straight road, we imported a suitable realistic road network based on OSM data, whereby the different speed limits were implemented by temporarily adding a random process in the data enhancement routine which drawed the speed limit $v^{max}_i$ for an edge $i\in E$ uniformly from $\{5, 6, 7, 8, 9\}$.
For all the properties of the vehicle and for all the remaining options of the software we used the default values.
Furthermore, we decided that the agent can perceive the vehicles' current velocity $v(t)\:[\frac{m}{s}]$ and the speed limit on its current edge $v^{max}_i$ with $i\in E$.\\\\
For the training of the agent we used the DDPG algorithm \cite{ddpg} with the actor and critic represented by two neural networks each having three fully-connected hidden layers with 400, 300, and 200 units, respectively.
As the activation functions, we chose \textit{LeakyReLU} \cite{leaky-relu} with a slope coefficient of 0.3 for all the layers except for the output layer of the actor and the critic where a \textit{tanh} and a \textit{linear} function was used, respectively.
In both networks, all the weights were randomly initialized based on $\mathcal{N}(0, 0.05)$ and we used the mean squared error as the loss function as well as the Adam optimizer \cite{adam} for the training of both networks.
The learning rates were set to 0.00005 for the actor and to 0.001 for the critic as well as to 0.01 for the training of their target networks.\footnote{Please see \cite{ddpg} for more information about target networks.}
Furthermore, the batch size was set to 32 and the size of the replay buffer \cite{rb} as well as the warm-up time was set to 10,000. % training each iteration.
For the exploration of the agent, the following second-order autoregressive process was used, whereby the rate of exploration (initialized to 0.99995) decreased with a factor of 0.99995 per iteration starting from 40,000:
\begin{equation}
a(t) = 0.29 \cdot a(t-1) + 0.7 \cdot a(t-2) + \mathcal{N}(0, 0.05)
\end{equation}
For the calculation of the reward $r$ of the agent we used
\begin{equation}
r(t) = \exp \Big (- 0.5 \cdot \Big (\frac {v^{max}_i - v(t)} {2.5} \Big )^2 \Big ) - 1,
\end{equation}
where again $v^{max}_i$ represents the speed limit of the vehicles' current edge $i\in E$.
Hence, the reward function was represented by a gaussian function producing maximum reward at the point where the vehicles' velocity equaled the prevailing speed limit.\\\\ 
Based on this setting, the task was handled after about 120,000 time steps corresponding approximately to a training time of 45 minutes on a Intel Core i5 without GPU acceleration. 
Figure 7a shows the smoothed average reward of the agent over 10 instances of this setting.
Furthermore, figure 7b presents one episode of the agent after training and it can be seen that the agent learned to use its actions precisely to accelerate and brake the vehicle towards different speed limits.

\begin{figure*}[ht!]
  \begin{tabularx}{\linewidth}[t]{*{1}X}
\setlength{\tabcolsep}{0.5em} % for the horizontal padding
{\renewcommand{\arraystretch}{0.1}% for the vertical padding
    \begin{tabular}[c]{p{\linewidth}}
\centering
     \subfigure[Smoothed average reward]{\includegraphics[scale=0.3]{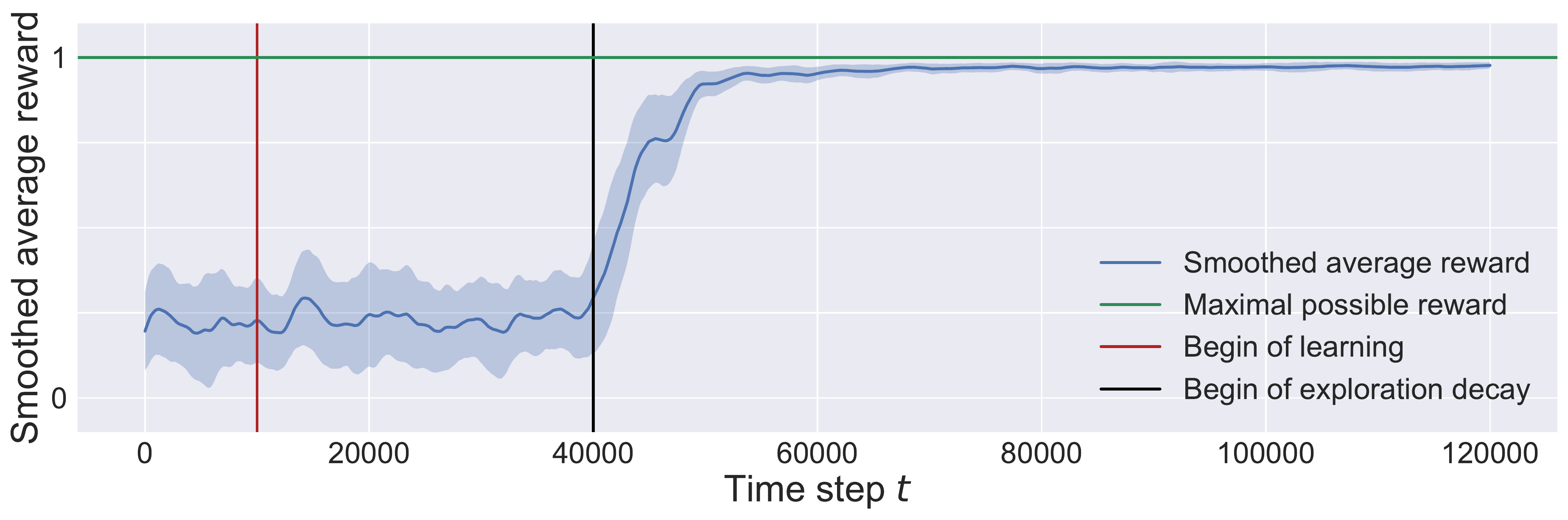} \label{f:ddpg}}
    \end{tabular}}
\tabularnewline
\setlength{\tabcolsep}{0.5em} % for the horizontal padding
{\renewcommand{\arraystretch}{0.1}% for the vertical padding
    \begin{tabular}[c]{p{\linewidth}}
 \centering
      \subfigure[Sample episode after training]{\includegraphics[scale=0.3]{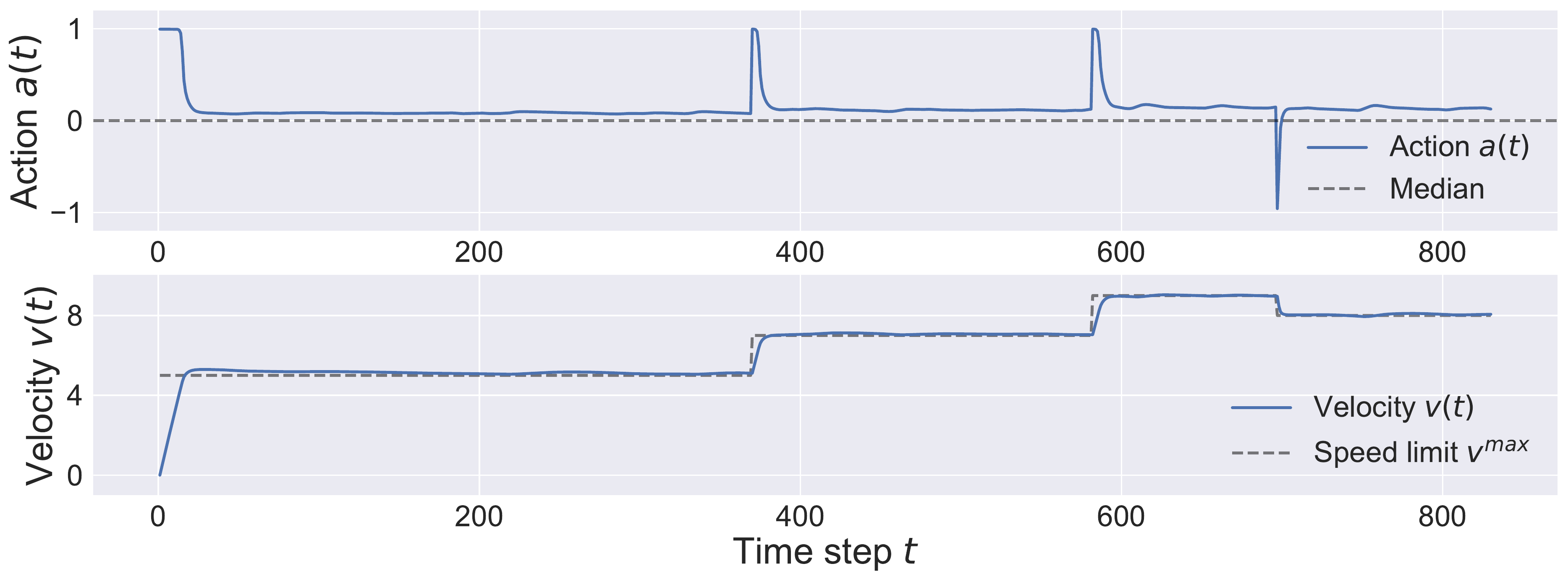} \label{f:ddpg-values}}
    \end{tabular}}
\end{tabularx}
\end{figure*}

\end{document}